\documentclass{article}

\PassOptionsToPackage{numbers, compress}{natbib}

\usepackage[preprint]{neurips_2022}

\usepackage[utf8]{inputenc} 
\usepackage[T1]{fontenc}    
\usepackage{hyperref}       
\usepackage{url}            
\usepackage{booktabs}       
\usepackage{amsfonts}       
\usepackage{nicefrac}       
\usepackage{microtype}      
\usepackage{xcolor}         

\usepackage{graphicx}

\usepackage{amsmath}

\usepackage{algorithm,algorithmic}

\usepackage{tabularx}

\usepackage{hhline}

\usepackage{adjustbox}

\usepackage{multirow}
\usepackage{enumitem}

\title{TKIL: Tangent Kernel Approach for Class Balanced Incremental Learning}

\author{%
  Jinlin Xiang\\
  Department of Electrical \&
  Computer Engineering\\
  University of Washington\\
  Seattle, WA 98195  \\
  \texttt{jinlinx@uw.edu} \\
  \And
  Eli Shlizerman\\
  Department of Electrical \& 
  Computer Engineering\\ Department of Applied Mathematics\\
  University of Washington\\
  Seattle, WA 98195  \\
  \texttt{shlizee@uw.edu} 
}

\begin{document}

\maketitle

\begin{figure}[ht]
\centering
\includegraphics[width=\linewidth]{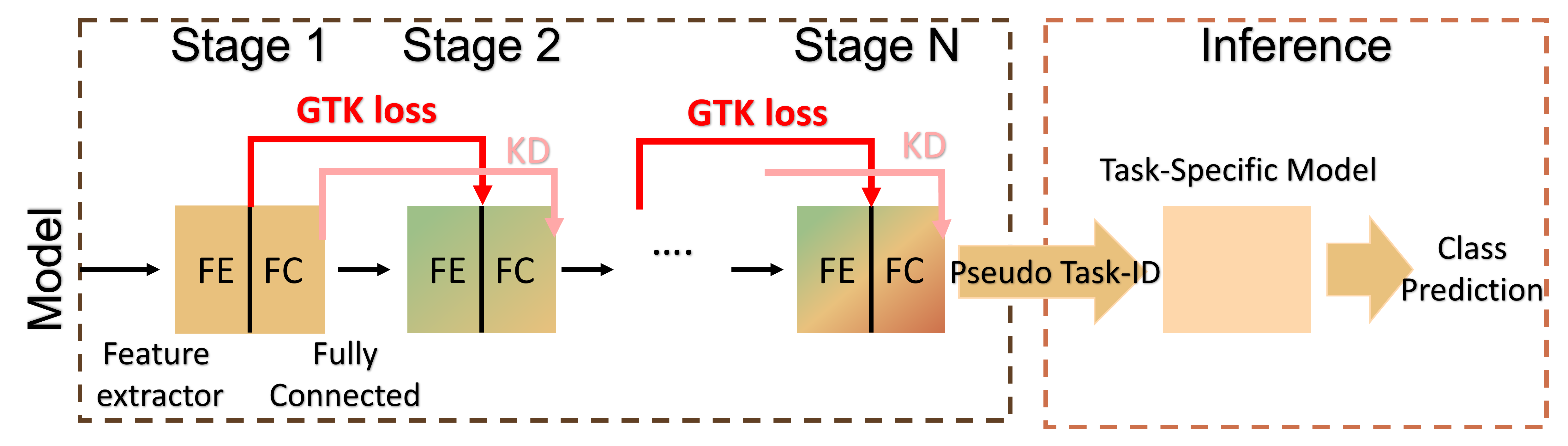}
\caption{Illustration of Tangent Kernel Approach for class balanced Incremental Learning (TKIL) , which implements a novel Gradients Tangent Kernel (GTK) loss in training contributing to optimal class balanced learning.} 
\label{fig:Overview}
\end{figure}

\begin{abstract}
When learning new tasks in a sequential manner, deep neural networks tend to forget tasks that they previously learned, a phenomenon called catastrophic forgetting. Class incremental learning methods aim to address this problem by keeping a memory of a few exemplars from previously learned tasks, and distilling knowledge from them. However, existing methods struggle to balance the performance across classes since they typically overfit the model to the latest task. In our work, we propose to address these challenges with the introduction of a novel methodology of Tangent Kernel for Incremental Learning (TKIL) that achieves class-balanced performance. The approach preserves the representations across classes and balances the accuracy for each class, and as such achieves better overall accuracy and variance. TKIL approach is based on Neural Tangent Kernel (NTK), which describes the convergence behavior of neural networks as a kernel function in the limit of infinite width. In TKIL, the gradients between feature layers are treated as the distance between the representations of these layers and can be defined as Gradients Tangent Kernel loss (GTK loss) such that it is minimized along with averaging weights. This allows TKIL to automatically identify the task and to quickly adapt to it during inference. Experiments on CIFAR-100 and ImageNet datasets with various incremental learning settings show that these strategies allow TKIL to outperform existing state-of-the-art methods.
\end{abstract}

\section{Introduction}
Deep Neural Networks (DNN) extensively advanced image processing applications, such as image classification, segmentation, recognition, and tracking~\cite{he2016deep,long2015fully,nam2016learning,noh2015learning}. However, their state-of-the-art performance is typically achieved on offline problems. When it comes to training models in a sequential manner, the accuracy drops significantly~\cite{french1999catastrophic,goodfellow2013empirical,kemker2018measuring}. The problem occurs primarily when DNNs confront new incoming tasks. For such a scenario, DNNs tend to override previously obtained knowledge and overfit to the latest task~\cite{masana2020class}. 

Class incremental learning aims to address this problem by keeping exemplars from previously learned tasks~\cite{rebuffi2017icarl,van2019three}. Due to privacy issues or storage constraints, only a limited number of exemplars from previous tasks can be kept. Advanced incremental learning methods address memory management and extraction of information from it in different ways, e.g., employment of a finetuning phase to re-learn previous tasks and to avoid overfitting to the latest task~\cite{li2017learning,castro2018end,rajasegaran2020itaml}. While finetuning is an effective way to balance classes, it assumes that the task, to which the classes belong, can be determined with high-fidelity before finetuning. Only a few existing methods incorporate task prediction during finetuning, although the prediction of the task is key in finetuning the base model to a corresponding task-specific model during inference~\cite{rajasegaran2020itaml,rebuffi2017icarl,wu2019large}.  Indeed, class incremental learning experiments show that a more accurate task-specific model can easily outperform the base model~\cite{rajasegaran2020itaml}.

The bias with respect to the latest task and the difficulty in predicting the associated task in class incremental learning was linked with diverging feature representations, i.e., the feature representations of each task-specific model would be very different from each other~\cite{kang2022class,aljundi2019gradient,wu2019large,hu2021distilling}. There is therefore a need to retain and organize the task feature representations, in an unbiased way to the order that the tasks were learned, such that they will be easily retrievable in the inference stage.  

Given these constraints, we thus propose to design and train an unbiased base model, capable to generate accurate task predictions, which in inference, translate to more effective finetuning to a specific task. To organize and retain the feature representations across tasks, we propose a novel Tangent Kernel approach for Incremental Learning (TKIL). It is inspired by Neural Tangent Kernels (NTK), which capture by constant tangent kernels the training dynamics that occur as a result of the gradient flow~\cite{jacot2018neural}. NTK were shown to converge to gradient-descent based training of ultra-wide (over-parameterized) DNN. While the scenario of ultra-wide networks does not apply to class incremental learning, the concept of capturing the gradient flow through tangent kernels is imperative for organizing feature representations learned during incremental training. Indeed, since gradient descent is being employed for optimizing each incremental step, capturing the gradient flow between the models used in previous and current learning steps would assist in coordinating the corresponding feature representations and prevent their divergence from each other. Specifically, we propose to compute the distance between the gradients of previous and current feature presentations through Gradients Tangent Kernels (GTK) and to minimize this distance as a loss in the optimization of class incremental learning. While GTK loss improves correspondence between feature representations, we find that practically, standard optimization could still cause overfitting to the current task due to imbalances between classes in the memory. To avoid such scenarios, we thereby propose a novel and appropriate training strategy for training task-specific models, i.e., average weights update rule. The rule trains each task model in a single mini-batch and then averages the weights of the model.

In summary, in this work, we propose a novel class balanced incremental learning method, TKIL, which includes a novel loss function, GTK Loss, for class balanced incremental learning. Optimizing GTK loss along with the average weights update rule, allows incremental learning to preserve the knowledge obtained on already learned tasks by balancing and achieving unbiased feature representations. Such balanced representations allow TKIL to generate robust task predictions during inference and successfully finetune to corresponding task-specific models. Experiments on MNIST, SVHN, CIFAR-100, and ImageNet show that TKIL achieves high accuracy (nearly 100\% accuracy) on task predictions and this accuracy translates to outperforming existing incremental learning methods, especially in multi-class incremental learning scenarios.

\section{Related Work}
\textit{\textbf{Class Incremental Learning.}} Class incremental learning is composed of three components: exemplars selection with rehearsal, forgetting constraints, and bias correction, as outlined in iCaRL approach~\cite{rebuffi2017icarl}. In addition, End-to-end IL~\cite{castro2018end} proposed the finetuning phase to boost the overall performance. Successive works contributed to enhancement of some of these four components. We describe in detail the components blow.

\begin{itemize}[leftmargin=*]
    \item \textit{\textbf{Rehearsal with exemplars.}} Rehearsal methods store a small set of previous task examples in a memory buffer to represent the whole task. Multiple previous works use herding heuristics~\cite{welling2009herding} to select the most representative exemplars ~\cite{rebuffi2017icarl,castro2018end,hou2019learning}. In addition, some works estimated the distribution of previously learned tasks and generated extra pseudo exemplars or images to avoid the imbalance between classes~\cite{liu2020mnemonics,ostapenko2019learning,odena2017conditional}. For example, CIS Gans~\cite{odena2017conditional} deployed a class conditional image synthesis network to generate fake images. In our approach, there is no need for complex methods to boost the classification accuracy, and we simply sample from each class randomly.
\item \textit{\textbf{Forgetting constraints.}}
Several works used regularization terms with classification loss as forgetting constraints ~\cite{kirkpatrick2017overcoming,li2020few,chaudhry2018riemannian,aljundi2018memory}. In addition, various works adopt Knowledge Distillation (KD) of class-IL to restore previous knowledge~\cite{rebuffi2017icarl,castro2018end,you2022incremental}. AFC~\cite{kang2022class} further modified the KD to define a discrepancy loss, which estimates the representation changes and retains the important representation knowledge. However, KD loss cannot constrain the divergence in feature representations and typically overfits the model to the latest task. We introduce the Gradients Tangent Kernel Loss to preserve representations. The GTK loss minimizes the divergences of feature representations between current and previous models and obtains balanced feature representations, which are key in obtaining an unbiased base model. As in other existing methods, we employ KD loss and classification loss to transfer the knowledge from the previous model.
\item \textit{\textbf{Bias-correction.}}
Various methods show that class-imbalance phenomena create a significant bias in class weights~\cite{zheng2020bi,hou2019learning}. Several works rectify this bias in different ways~\cite{wu2019large,belouadah2019il2m,hou2019learning,zheng2020bi}. For example, BIC corrects the last fully connected layer bias by introducing an extra linear model~\cite{wu2019large}. Incremental Dual memory employs a dual memory in data augmentation and rectifies the final activation functions to correct predictions~\cite{belouadah2019il2m}. The limitation of the above methods is that they only rectify the bias at the last linear layer. However, class-imbalance issue typically causes the whole model to overfit to the latest task. Our method optimizes the GTK loss with the average weights update rule, which rectifies the biases in all model layers, not only the last linear layer.
\item \textit{\textbf{Finetuning.}}
Finetuning with exemplars is an effective way to balance the bias and was shown to outperform well in the multi-class incremental learning scenarios. LwF~\cite{li2017learning} introduced a finetuning adaption loss while training a new task. End-to-end IL~\cite{castro2018end} finetuned the model by re-learning samples in the memory after learning a new task. iTAML~\cite{rajasegaran2020itaml} introduced an automatic task prediction method and finetuning phase in inference. However, the automatic task prediction doesn't perform well when the base model is biased. Our work adopts the task prediction method and trains a balanced base model with it.
\end{itemize}

\textit{\textbf{Neural Tangent Kernels.}}
NTK were first introduced in \cite{jacot2018neural} and developed in subsequent works~\cite{arora2019exact,li2019enhanced,novak2019neural,wallace2021can}. NTK employ the gradient of kernels to describe the convergence behavior and to mimic the performance of (over-parameterized) DNN within a limit of infinite width. TTK~\cite{wallace2021can} uses a kernelized distance across the gradients of multiple random initialized networks to estimate the similarity over different tasks. While the ultra-wide networks do not apply to incremental learning, capturing the gradient flow in each incremental learning step could assist in coordinating the corresponding feature representations. Inspired by NTK, we formulate a kernel gradients loss, GTK loss. Such loss minimizes the gradients across current and previous models to prevent their divergence from each other.

\section{Method}
\subsection{Problem Setup}
\begin{figure}[t]
\centering
\includegraphics[width=\linewidth]{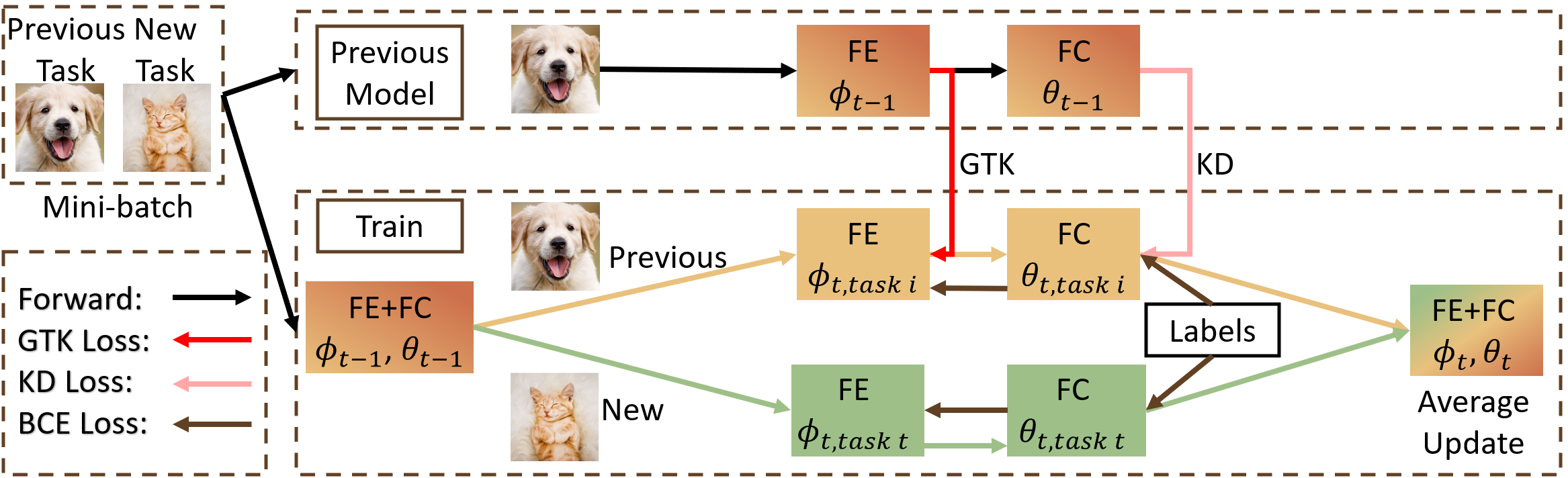}
\caption{Illustration of Training with GTK loss. For the previous tasks (task i, i = 1,2,3,...t-1), the current model optimizes the Binary cross entropy loss between labels and predictions (brown color), knowledge Distillation loss between previous model soft predictions and current model predictions (pink color), and the Gradients Tangent kernel loss between previous model feature extractor gradients map and current model feature extractor gradients map (red loss) over a mini-batch sampled from memory and new data sets. The new task(task t), only employs Binary Cross Entropy loss (brown loss backward) (FE: Feature extractor layers, FC: Fully Connected layers)} 
\label{fig:model}
\end{figure}

In class incremental learning, the objective is to learn a unified classifier in a sequential manner. We demonstrate our training approach in Figure~\ref{fig:model} and inference approach in Figure~\ref{fig:inference}. More formally, we denote the sequence of data as a batch of class sets  $\mathcal{D}\!=\!\{D_{1},D_{2},\dots,D_{t},\dots,D_{N}\}$ of $N$ tasks, where $D_{t}$ contains the training images $X\!=\!\{x_j\}_{j=1}^{n}$ and labels $Y\!=\!\{y_j\}_{j=1}^{n}$, where~${x}_{j}$ is one image input, and ${y}_{j}$ is one ground-truth label, $n$ is number of images. At the $t$-th incremental stage, complete data for new classes $D_{t}$ and a small set of previous exemplars~$M_{t}\!=\!\{\mathcal{M}_1,\dots,\mathcal{M}_{t-1}\}$ in fixed memory buffer is available, where~$\mathcal{M}_1 \in D_{1},\dots,\mathcal{M}_{t-1} \in D_{t-1}$. The model is expected to classify all the classes seen so far. We denote the $t$-th expert model as $F_t(x)\!=\!F(\phi_{t},\theta_{t}, x)\!=\!f_t(g_t(x))\!=\!f(\theta_{t},g(\phi_{t}, x))$, including feature extractor layers $g$ with weights $\phi_{t}$ and fully connected layers~$f$ with weights $\theta_{t}$. During the training, we additionally employ previous weights, $\phi_{t-1}, \theta_{t-1}$, as the teacher weights. In the $t$-th incremental training stage, we train the expert model to minimize all three losses (Class Loss, KD Loss, and GTK Loss) with previous exemplars $M_{t}$ and current dataset $D_{t}$. 

Class Loss is computed as the distance between expert model predictions~$F_t(x)$ and ground truth labels~$Y$; Knowledge distillation loss is between expert model predictions~$F_t(x)$ and soft labels~$F_{t-1}(x)$; The GTK Loss is between the gradient of expert model feature representations~$G_{t}(x)$ and the gradient of previous model feature representations~$G_{t-1}(x)$. To avoid overfitting to the latest task during training, we apply an average weights updating rule. In each mini-batch, we optimize corresponding task-specific models by above three losses, and then average the weights of all task-specific models ~$\{\phi_{t,\textit{task }i},\theta_{t,\textit{task} i}, i =1,2,\dots,t\}$ to be the current model ($F_t$). Based on the setting above, we define the TKIL objective below. This approach is shown in Figure~\ref{fig:model}.

\textit{\textbf{Objective of TKIL.}} \textit{At the current site $t$, our goal is to continuously optimize an expert base model $F_t(\phi_t, \theta_t)$ along with $\mathcal{L_\textit{Class}}$, $\mathcal{L_\textit{KD}}$ and $\mathcal{L_\textit{GTK}}$ based on the knowledge from memory~$M_t$ and new task$D_t$}. This model can (1) generate an accurate task prediction, and (2) corresponding task-specific model ($F_{\textit{task}}$) achieves competitive performance on the class prediction during the inference stage.

\subsection{Gradients Tangent Kernel}
\begin{figure}[ht]
\centering
\includegraphics[width=\linewidth]{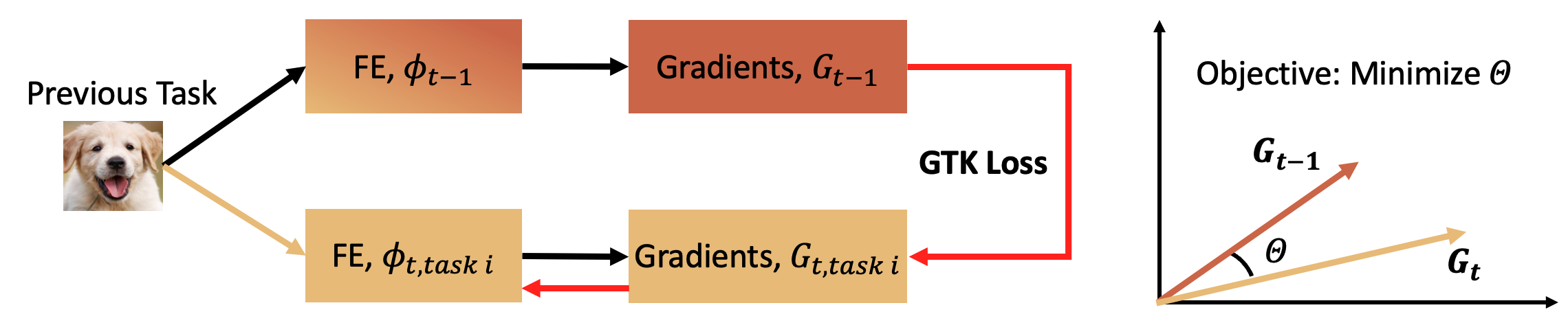}
\caption{Illustration of Gradients Tangent Kennel Loss, which estimates the cosine distance of two gradients feature presentations and minimizes this kernel cosine distance over a BCE loss.} 
\label{fig:loss}
\end{figure}

Deep neural networks, such as ResNet-18\cite{he2016deep}, usually contain Feature Extractors layers (FE), denoted as $g$ with weights $\phi$, and Fully Connected layers (FC), denoted as $f$ with weights $\theta$. The prediction of the deep neural networks at $t$-th incremental learning stage is expressed as
\begin{equation}
    \widehat{y} ={f(\theta_{t},h)} = {f(\theta_{t},g(\phi_{t}, x))},
\end{equation}

where $x$ is the input images and $\widehat{y} \in \mathbb{R}^{c}$($c$ is number of classes) is the predicted logits. $h$ is the output of the feature layers.

To organize and retain the feature representations across tasks, we propose the Gradients Tangent Kernel loss, as shown in Figure~\ref{fig:loss}. The TKIL approach is inspired by the Neural Tangent Kernels (NTK), which are defined as
\begin{equation}
    \mathbf{K}(x,x') = \mathbf{E}_{\theta}\left\langle\frac{\partial f(x,\theta)}{\partial \theta}, \frac{\partial f(x',\theta)}{\partial \theta}\right\rangle,
    \label{eq:ntk}
\end{equation}
where $x$ and $x'$ are the input data points and $\theta$ is the parameter of neural networks, typically drawn from a Gaussian distribution.

Our work adapts Neural Tangent Kernel in Eq.~\ref{eq:ntk} to a kernel loss, Gradient Neural Tangent Kernel loss (GTK loss), between the current and previous model. This kernel loss retains the gradient representation from diverging to the latest task by minimizing the cosine distance, $\Theta$, between the two gradients, $G_{t-1}$ and $G_t$ respectively, which also shown in Figure~\ref{fig:loss}. The gradients are the derivative of loss with respect to the last feature layer from the current task model with the weights $\phi_{\textit{t}}$ and previous expert models with weights $\phi_{t-1}$. We define the gradients and objective function in following
\begin{align}
    &G_{t-1} = \frac{\partial g(\phi_{t-1},x)}{\partial \phi_{t-1}}, G_{t} = \frac{\partial g(\phi_{t},x)}{\partial \phi_{t} },\\
    &\min_{F_t} \mathbb{E}_{(x,y) \sim {M_t}}[\mathcal{L}_{GTK}(\frac{\left\langle G_{t}, G_{t-1} \right\rangle}{\|G_{t}\|\|G_{t-1}\|})],
\end{align}
where $G_{t-1}$ and $G_{t}$ are the gradients of the last feature layer from the previous and current model respectively. Noted that the $G_{t}$ is gradients of one task model $F_{t, \textit{task i}}$ in practice. $M_t$ is the exemplars from previously learned classes in the memory. We employ a Binary Cross-Entropy (BCE) loss to minimize the distance $\Theta$.

\subsection{Averaging Weights Training Approach}

To avoid overfitting to the current task, we propose the rule of average weights update in one mini-batch. We separate data samplers into different tasks and train the task-specific models (where weights are $\phi_{t,\textit{task }i}$ and $\theta_{t,\textit{task }i}$) corresponding to different tasks. At the end of the mini-batch, we compute the average of all weights as final weights ($\phi_t$ and $\theta_t$). This rule rectifies the whole model, not only the last linear layers. This training approach is also illustrated in Figure~\ref{fig:model} and as pseudo code Algorithm~\ref{tab:algorithm}.

More formally, we demonstrate this approach with our objective function in Algorithm~\ref{tab:algorithm}. In the mini-batch at $t$-th incremental stage, we have a union batch from previous exemplars~$M_t$ and current dataset~$D_t$, noted $B_t \sim {D_t \cup M_t}$. Training starts from the previous expert model~$F_{t-1}$ with wights $\phi_{t-1},\theta_{t-1}$. We initialize the current task model with same weights ($\phi_{t-1},\theta_{t-1}$) from the previous expert model. 

For all samples, we employ the classification loss to minimize the loss between predicted logits~$\widehat{y}$ and labels~$y$, which is expressed as
\begin{equation}
    \min_{F_t} \mathbb{E}_{(x,y) \sim {D_t \cup M_t}}[\mathcal{L_{\textit{Class}}}(\widehat{y},y)],
\end{equation}
where ${D_t \cup M_t}$ is the data distribution of the previous tasks(task 1 to task $t-1$) and new task~$t$. $\mathcal{L}$ is denoted as the task-specific loss, e.g., binary cross-entropy classification loss in our case. 

For samples from previously learned tasks, we employ Knowledge Distillation (KD) Loss as a forgetting constraint term. KD Loss penalizes the change with respect to the output from the previous expert model by a BCE loss. The KD objective function is expressed as
\begin{equation}
    \min_{F_t} \mathbb{E}_{(x,y) \sim {M_t}}[\mathcal{\mathcal{L}_{\textit{KD}}}(F_{t-1}(x),F_{t}(x))]],
\end{equation}
where $F_t$ is the current expert model and $F_{t-1}$ is the previous expert model. For previous tasks~$(x,y) \sim {M_t}$, we minimize the KD loss. $F_{t-1}(x)$ is the soft prediction and $F_{t}(x)$ is the prediction of the current model. The KD loss then minimizes the difference between soft  and current predictions.

For the last step, we employ the GTK loss to constrain the divergence from the optimal feature representation. Thus the overall objective function contains three loss functions (classification loss, KD loss, and GTK loss), and is now expressed as ($\gamma$ is the hyperparameter GTK loss. $\alpha$ and  $\beta$ are hyperparameters in KD loss.):
\begin{align}\left\{\begin{aligned}
&\min_{F_t} \mathbb{E}_{(x,y) \sim {M_t}}[\alpha \mathcal{L}(F_{t}(x),\widehat{y}) + \beta\mathcal{L}(F_{t-1}(x),F_{t}(x)) +\gamma\mathcal{L}(\frac{\left\langle G_{t}, G_{t-1} \right\rangle}{\|G_{t-1}\|\|G_{t}\|}]\\
&\min_{F_t} \mathbb{E}_{(x,y) \sim {D_t}}[\mathcal{L}(F_{t}(x),\widehat{y})]
\end{aligned}\right.,\end{align}

We obtain the current expert model $F_{t}$ in a single mini-batch and average the weights according to weights averaging rule updating update rule:
\begin{equation}
    F_t = \frac{1}{t}\sum_i^t F_{t, \textit{task } i},
\end{equation}
where $F_{t, task i}$ is the task specific-model and the $F_t$ is the current model.

\begin{algorithm}[t]
\caption{TKIL Algorithm in one mini-batch}
\label{tab:algorithm}
\begin{algorithmic}[1]
    \REQUIRE Dataset: $D_t$, Memory: $M_t$, Note $B_t \in D_t \cup M_t$ Hyperparameters: $\alpha, \beta, \gamma$
    \STATE Initialize the Model (For the first mini-batch):
    \STATE \quad $F_t(\phi_t,\theta_t)$: $\textit{Previous Expert Model: } F_{t-1}(\phi_{t-1},\theta_{t-1}) \longrightarrow F_t(\phi_t,\theta_t)$
    
        \FOR {All training Sample in $B_t$}
            \FOR {$i$ = 1,2,3,..t }
            \STATE $F_t \longrightarrow F_{t, \textit{task } i }$
            \IF{i < t}
                \STATE $\mathcal{L}_{\text{Class}} =\sum_j^{\textit{task } i} \mathcal{L}_{\text{BCE}}(F_{t,\textit{task } i}(\phi_{t,\textit{task } i},\theta_{t,\textit{task } i},B_{t,j}), Y_j) $
                \STATE $\mathcal{L}_{\text{KD}} =\sum_j^{\textit{task } i} \mathcal{L}_{\text{BCE}}(F_{t,\textit{task } i}(\phi_{t,\textit{task } i},\theta_{t,\textit{task } i},B_{t,j}),F_{t-1}(\phi_{t-1},\theta_{t-1},B_{t,j})) $
                \STATE $\mathcal{L}_{\text{GTKL}} =\sum_j^{\textit{task } i} \mathcal{L}_{\text{BCE}}(G_{t,\textit{task } i}(\phi_{t,\textit{task } i},B_{t,j}), G_{t-1}(\phi_{t-1},B_{t,j})) $
                \STATE $\mathcal{L}_{\text{all}} = \alpha\mathcal{L}_{\text{Class}} + \beta\mathcal{L}_{\text{KD}} + \gamma\mathcal{L}_{\text{GTKL}} $
                \STATE $F_{t,\textit{task } i}= (\phi_{t,\textit{task } i},\theta_{t,\textit{task } i} )$ by minimizing the $\mathcal{L}_{\text{all}}$
        
            \ELSIF{i = t}
            \STATE $\mathcal{L}_{\text{Class}} =\sum_j^{\textit{task } i} \mathcal{L}_{\text{BCE}}(F_{t,\textit{task } i}(\phi_{t,\textit{task } i},\theta_{t,\textit{task } i},B_{t,j}), Y_j) $
              \STATE $F_{t,\textit{task } t}= (\phi_t,\theta_t )$ by minimizing the $\mathcal{L}_{\text{class}}$
              \ENDIF
             \ENDFOR
             \STATE $F_t \longleftarrow \frac{1}{t}\sum_i^t F_{t, \textit{task } i}$
        \ENDFOR
\end{algorithmic}

\end{algorithm}

\subsection{Inference}
\begin{figure}[ht]
\centering
\includegraphics[width=\linewidth]{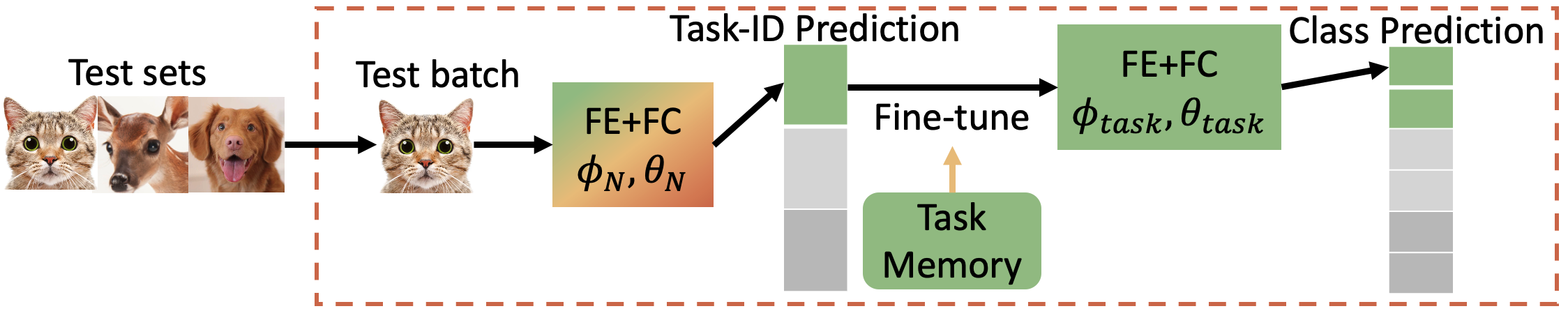}
\caption{In Inference, the base model generates task predictions when given the test sets for all samples. Then the predicted corresponding task model  with highest probability is finetuned from base model with exemplars from memory to predict the class.} 
\label{fig:inference}
\end{figure}
During the inference stage, we adopt the task prediction procedure described in iTAML~\cite{rajasegaran2020itaml}. As we have mentioned, the finetuned task model ($F(\phi_{\textit{task}}$, $\theta_{\textit{task}})$) can easily surpass the base model ($F(\phi_N,\theta_N)$) when given an accurate task prediction. Thus, training a base model to generate an accurate task Pseudo ID is important. As this process is purely self-supervised, we can only improve the base model task performance during the training. The GTK loss and the average training approach ensure we obtain an unbiasd base model. The inference pipeline is shown in Figure~\ref{fig:inference}.


\section{Experiments and Results}

\subsection{Datasets and Implementation }\label{setting}
We conduct experiments on four popular benchmark, i.e., CIFAR100~\cite{krizhevsky2009learning} and ImageNet~\cite{russakovsky2015imagenet} (ImageNet-100 and ImageNet-1k). CIFAR100 dataset contains $50,000$ training images and $10,000$ testing images. ImageNet-1k dataset consists of $1,281,167$ images for training and $50,000$ images for validation across $1,000$ classes. ImageNet-100 dataset includes 100 randomly sampled classes from ImageNet-1k.

We employ a standard ResNet-18~\cite{he2016deep} as our base model. For all datasets, we use RAdam~\cite{liu2019variance} with an initial learning rate of 0.01 for 70 epochs (which is notable less than the previous method, like PODNet~\cite{douillard2020podnet} needs 160 epochs), and a batch size of 128 for CIFAR, and 80 for ImageNet. The learning rate is multiplied by 10 after every 20 epochs. All models are trained on two 2080Ti GPUs with parallel computation mode. The proposed method is implemented based on the publicly available official code~\cite{rajasegaran2020itaml,javed2018revisiting}. For fair comparisons with previous methods, we report average incremental accuracy over seen classes across all incremental stages with the same setting. We select iCaRL as the baseline method. 
We don't include the methods, like LwF~\cite{li2017learning} and EWC\cite{kirkpatrick2017overcoming}, which achieve lower performance than iCaRL.
\begin{table*}[ht]
	\begin{center}
	\caption{ Performance comparison between the proposed algorithm and other state-of-the-art methods on CIFAR100, ImageNet-100 and ImageNet dataset.}
	\label{tab:results}
    \begin{adjustbox}{width=0.9\linewidth}
	\begin{tabular}{ccccccc}\toprule
	\textbf{Method} & \multicolumn{2}{c}{\textbf{CIFAR-100}} & \multicolumn{2}{c}{\textbf{ImageNet-100}} & \multicolumn{2}{c}{\textbf{ImageNet-1k}} \\
	 \cmidrule(r){2-3} \cmidrule(r){4-5} \cmidrule(r){6-7}
	\textbf{Number of Tasks} &5&10& 5&10& 5&10\\
	\midrule
 	iCaRL~\cite{rebuffi2017icarl}& 57.2\% &52.6\% &65.1\%& 59.6\% &51.5\% & 46.8\% \\
 	BIC~\cite{wu2019large}& 59.3\% &54.2\% &70.1\%& 64.9\% &62.6\% & 58.7\% \\
 	DDE~\cite{hu2021distilling}&59.1\% & 55.31\% & 69.2\%& 65.5\% &- & - \\
 	Mnemonics~\cite{liu2020mnemonics} &63.3\% &62.2\% &72.6\%& 71.4\% &64.5\% & 63.5\% \\
 	
 	PODNet~\cite{douillard2020podnet}&- &59.7\% &72.5\% &71.5\% &64.1\% & 62.0\% \\

 	iTAML~\cite{rajasegaran2020itaml}&74.5\% &76.6\% & 69.3\% &70.4\%& 64.4\% &63.4\% \\
 	DER~\cite{yan2021dynamically}&76.8\% & 75.4\% & - & -&- & - \\
 	\midrule
 \textbf{TKIL(Ours)}&\textbf{79.5\%} &\textbf{82.5\%} &\textbf{73.7\%} &\textbf{73.3\%}& \textbf{65.9\%} &\textbf{64.8\%}\\
 	
	\bottomrule
	\end{tabular}
    \end{adjustbox}
    \end{center}
\end{table*}
\subsection{Experimental Results}
We test TKIL on various incremental learning scenarios and compare it with existing methods on CIFAR-100, ImageNet-100, and ImageNet-1k. We show evaluation results for incremental learning of 5 and 10 tasks in Table~\ref{tab:results}. The results indicate that TKIL achieves accuracy that is more optimal than other compared methods on all compared scenarios. The margin in the accuracy of TKIL vs. existing approaches is particularly evident in the 10 tasks (stages) scenario which is consistent with the intent of TKIL to balance performance across tasks and classes.
For example, on CIFAR-100 with 10 incremental stages, TKIL achieves an improvement of 5.9\% in accuracy vs. iTAML. Indeed, as the number of stages grows, the gap in accuracy of TKIL with respect to existing methods is expected to grow even further. To test this postulation, we consider the CIFAR-100 dataset with even further stages of incremental learning, i.e., 25 tasks where 2 classes are learned each time, and show the results in Table~\ref{tab:20results}. In such a scenario, the accuracy of existing state-of-the-art method, iTAML, drops to 55.87\% from 76.6\% due to unstable prediction of tasks and AFC becomes a leading existing method with 64.34\%. On the other hand, TKIL approximately preserves its performance achieving accuracy of 78.97\% which constitutes a gap of 14.63\% in accuracy between TKIL and the second best performing method AFC. These results indicate that KD alone is insufficient to address multi-class and multi-stages scenarios of incremental learning and additional components, such as the components we introduce in TKIL, are needed.  In addition, we conduct similar experiments on MNIST\cite{deng2012mnist} and SVHN~\cite{netzer2011reading} (included in Supplementary Materials). TKIL achieves 97\% or more on these benchmarks when learning 2 classes each time. These experiments show that TKIL predicts consistently stable and accurate task predictions and as a result it translates to more optimal accuracy for multi-class problems.
\begin{table*}[h]
	\begin{center}
	\caption{ Performance comparison between the proposed algorithm and other state-of-the-art methods on CIFAR-100 dataset, 25 stages with 2 classes each stage.}
	\label{tab:20results}
    \begin{adjustbox}{width=\linewidth}
	\begin{tabular}{ccccccc}\toprule
	\textbf{Method} & iCaRL~\cite{rebuffi2017icarl} & 	BIC~\cite{wu2019large} &iTAML~\cite{rajasegaran2020itaml} & PODNet~\cite{douillard2020podnet} &AFC\cite{kang2022class} &  TKIL(ours)\\ \midrule
 	\textbf{Accuracy} & 50.60\%  & 48.65\% & 55.87\% &63.31\% & 64.34\%&\textbf{78.97 (+14.63)\%}\\
	\bottomrule
	\end{tabular}
    \end{adjustbox}
    \end{center}
\end{table*}

\subsection{Analysis and Ablation Studies}\label{limitation}

\begin{figure}[t]
\centering
\includegraphics[width=\linewidth]{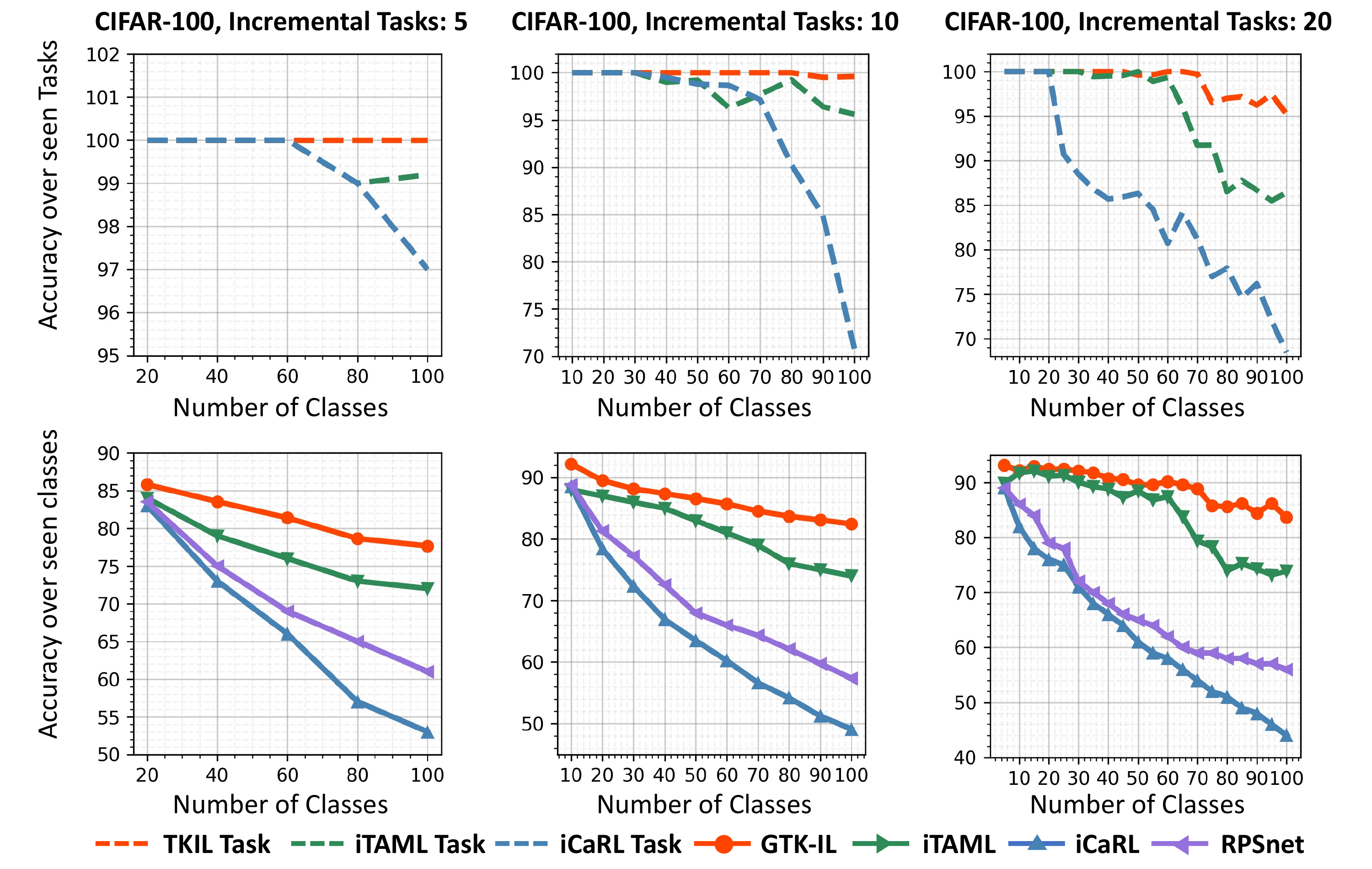}
\caption{Top: Task accuracy over seen classes on CIFAR-100, with 10, 20 and 5 tasks from left to right; bottom: classification accuracy over seen classes on CIFAR100, with 10, 20 and 5 tasks from left to right. TKIL consistently outperforms other existing methods across different settings.} 
\label{fig:ablation}
\end{figure}

\textbf{\textit{Task Prediction and Classification Accuracy at Different Stages.}} In Figure~\ref{fig:ablation} we investigate with CIFAR-100 benchmark the accuracy of the compared approaches when tasks are added incrementally, i.e., 5, or 10, or 20 classes are added each time. In these experiments, TKIL consistently outperforms the compared methods by a large margin (e.g. 83\% v.s. 73\% in 20 stages scenario). Indeed, these results demonstrate that while adding classes impacts overall accuracy of all methods, TKIL succeeds to maintain the smallest slope of the drop and no significant collapses events in the accuracy curve. In addition, we compute the accuracy of task prediction for all compared methods and add the prediction component to methods if they did not originally include one, such as iCaRL. We show the results in Supplementary Material that demonstrate that TKIL obtains almost perfect task prediction while other methods would display a decrease or even a drop in task prediction accuracy after a few stages. 

\begin{table*}[ht]
	\begin{center}
	\caption{Ablation study of hyperparameter $\gamma$ selection}
	\label{tab:hyperparameter}
    \begin{adjustbox}{width=0.65\linewidth}
	\begin{tabular}{cccccc}\toprule
 		\multirow{2}{*}{\textbf{Parameter}}& \textbf{S1} &\textbf{S2}&\textbf{S3}&\textbf{S4}&\textbf{S5}\\
 	\cmidrule(r){2-6}
 	&\multicolumn{5}{c}{\textbf{Task-prediction over all seen classes}}\\
 	\midrule
    $\gamma = 10$ & 100\%&82.0\%&33.5\%&21.9\%&17.5\% \\
    $\gamma = 1$ & 100\%&100\%&97.5\%&94.3\%&92.5\% \\
    $\gamma = 0.1$& 100\%&100\%&100\%&100\%&100\%\\
    $\gamma = 0$& 100\%&100\%&99.5\%&95.6\%&98.4\%\\
	\bottomrule
	\end{tabular}
    \end{adjustbox}
    \end{center}
\end{table*}
\textbf{\textit{Selection of hyperparameter $\gamma$.}} In Table~\ref{tab:hyperparameter} we investigate the impact of the GTK loss. We compare TKIL method with its variants with parameters ($\gamma = 0, 0.1, 1,10$) to estimate the ability of retaining the feature representations with it. $\gamma$ controls GTK loss and when selecting  $\gamma$ to be large, the model remembers the first task and could overfit to it and could fail to make task predictions. When letting $\gamma =$ 0, the model  forgets the optimal representations and diverges to the latest task, which leads to an unstable task-prediction. Based on these results we select $\gamma$ as $\gamma = 0.1$.

\begin{table*}[ht]
	\begin{center}
	\caption{Ablation study on TKIL components: KD, GTK Loss and Average Rule on CIFAR-100, 10 Incremental stages}
	\label{tab:abaltion}
    \begin{adjustbox}{width=\linewidth}
	\begin{tabular}{cccccccccccc}\toprule
 	Component&& Stage 1& Stage 2& Stage 3& Stage 4& Stage 5& Stage 6& Stage 7& Stage 8& Stage 9& Stage 10\\\midrule
 	\multirow{2}{*}{KD} &Task $\uparrow$ & 100\%&100\%&100\%&99.0\%&97.2\%&95.6\%&94.0\%&81.5\%&74.2\%&47.6\%\\
 	& Class $\uparrow$ & 90.7\%&88.0\%&86.6\%&82.8\%&79.7\%&76.8\%&74.6\%&62.8\%&57.8\%&46.6\%\\ \midrule
 	{KD + } &Task $\uparrow$ & 100\%&100\%&100\%&99.5\%&100\%&95.6\%&100\%&97.5\%&94.2\%&96.1\%\\
 	{Average Rule}& Class $\uparrow$ & 90.3\%&86.2\%&86.5\%&84.4\%&81.3\%&81.2\%&78.7\%&79.1\%&77.2\%&76.4\%\\ \midrule
 	\multirow{2}{*}{TKIL} &Task $\uparrow$ & 100\%&100\%&100\%&100\%&100\%&100\%&100\%&100\%&100\%&\textbf{99.9\%}\\
 	& Class $\uparrow$ & 92.2\%&89.6\%&88.2\%&87.4\%&86.6\%&85.7\%&84.6\%&83.7\%&83.2\%&\textbf{82.5\%}\\ 

	\bottomrule
	\end{tabular}
    \end{adjustbox}
    \end{center}
\end{table*}
\textbf{\textit{Components contribution.}} To demonstrate how different components contribute to TKIL performance, we investigated ablations of the TKIL architecture in Table~\ref{tab:abaltion}. We create baselines with KD only and  KD and Average Rule. We find that KD only task and class predictions decrease as more incremental stages are introduced. When considering KD with Average Rule, we observe improvement in task prediction from the KD only variant however these is still a decrease in accuracy for class prediction, eventually decreasing to 76.4\%. When GTK Loss is added we indeed observe a boost in accuracy on both task and class predictions with task prediction being 99.9\% and class prediction 82.46\%.

\textbf{\textit{Limitations.}} While TKIL approach provides improved and balanced accuracy for incremental learning, it might need additional resources than other methods. According to TKIL algorithm~\ref{tab:algorithm}, task-specific models need to be re-trained each time and gradients need to be computed for them. The offline training time is expected to increase to $t \times \textit{batch time}$ v.s. $\textit{batch time}$, where $t$ is the incremental stage. However, when a large enough training batch size can be set, this could mitigate the extra time spent on training time by using GPU computing.

\section{Conclusion}\label{con}
Here we propose a novel class-balanced incremental learning approach, TKIL, that aims to address both classes balanced accuracy and overall classification accuracy. TKIL trains a balanced base model that in inference exceeds the accuracy of state-of-art methods by predicting and adapting to a corresponding task-specific model. We show that the components of TKIL that support such performance are the GTK loss along with an average update, which is key for transferring and synchronizing knowledge between tasks and classes. Indeed, TKIL optimizes for retaining feature representations across stages of learning and rectifies the bias in the model. Our experiments on multiple benchmarks show that TKIL outperforms existing state-of-the-art methods on various incremental learning settings.




\bibliographystyle{unsrt}


\end{document}